\documentclass[journal, peerreview]{IEEEtran}
\pdfoutput=1
\usepackage{cite}
\usepackage[pdftex]{graphicx}
\DeclareGraphicsExtensions{.pdf}
\usepackage{amsmath}
\usepackage{amsfonts}
\ifCLASSOPTIONcompsoc
  \usepackage[caption=false,font=normalsize,labelfont=sf,textfont=sf]{subfig}
\else
  \usepackage[caption=false,font=footnotesize]{subfig}
\fi

\ifCLASSOPTIONcaptionsoff
  \usepackage[nomarkers]{endfloat}
 \let\MYoriglatexcaption\caption
 \renewcommand{\caption}[2][\relax]{\MYoriglatexcaption[#2]{#2}}
\fi

\usepackage{moreverb,url}
\usepackage[]{subfig}
\usepackage{ifthen}
\usepackage{textcomp,gensymb}
\usepackage{mathtools}
\usepackage[flushleft]{threeparttable}
\usepackage{array}
\usepackage{float}
\usepackage{import}
\graphicspath{{Figs/}}

\begin{document}

\title{Drivers’ Manoeuvre Modelling and Prediction for Safe HRI}

\author{Erwin~Jose Lopez~Pulgarin,~
        Guido Herrmann,~
        and~Ute Leonards
\thanks{E J. Lopez Pulgarin and G. Herrmann are with the Department of Electrical and Electronic Engineering (EEE), University of Manchester, Manchester, UK, (e-mail: erwin.lopezpulgarin@manchester.ac.uk;guido.herrmann@manchester.ac.uk).}
\thanks{U. Leonards is with the School of Psychological Science,University of Bristol, Bristol, UK, (e-mail: ute.leonards@bristol.ac.uk).}
}

\ifCLASSOPTIONpeerreview
	\markboth{IEEE TRANSACTIONS ON HUMAN-MACHINE SYSTEMS,~Vol.~N, No.~Y, August~20XX}%
 {IEEE TRANSACTIONS ON ROBOTICS,~Vol.~N, No.~Y, August~20XX}
\fi
	\markboth{IEEE TRANSACTIONS ON HUMANAN-MACHINE SYSTEMS,~Vol.~N, No.~Y, August~20XX}%
{LOPEZ PULGARIN \MakeLowercase{\textit{et al.}}: DRIVERS’ MANOEUVRE MODELLING AND PREDICTION FOR SAFE HRI}

\maketitle

\begin{abstract}
As autonomous machines such as robots and vehicles start performing tasks involving human users, ensuring a safe interaction between them becomes an important issue. Translating methods from human-robot interaction (HRI) studies to the interaction between humans and other highly complex machines (e.g. semi-autonomous vehicles) could help advance the use of those machines in scenarios requiring human interaction. One method involves understanding human intentions and decision-making to estimate the human's present and near-future actions whilst interacting with a robot. This idea originates from the psychological concept of Theory of Mind, which has been broadly explored for robotics and recently for autonomous and semi-autonomous vehicles. In this work, we explored how to predict human intentions before an action is performed by combining data from human-motion, vehicle-state and human inputs (e.g. steering wheel, pedals). A data-driven approach based on Recurrent Neural Network models was used to classify the current driving manoeuvre and to predict the future manoeuvre to be performed. A state-transition model was used with a fixed set of manoeuvres to label data recorded during the trials for real-time applications. Models were trained and tested using drivers of different seat preferences, driving expertise and arm-length; precision and recall metrics over 95$\%$ for manoeuvre identification and 86$\%$ for manoeuvre prediction were achieved, with prediction time-windows of up to 1 second for both known and unknown test subjects. Compared to our previous results, performance improved and manoeuvre prediction was possible for unknown test subjects without knowing the current manoeuvre.
\end{abstract}


\ifCLASSOPTIONpeerreview
\begin{center} \bfseries This work is part of the lead author's PhD studies \cite{pulgarin2019estimation}. This work has been submitted to the IEEE for possible publication. Copyright may be transferred without notice, after which this version may no longer be accessible. \end{center}
\fi

\IEEEpeerreviewmaketitle

\section{Introduction}
\label{sec:intro}

The field of robotics has kept expanding during the last decade, generating new technical and methodological advances. As these start to be integrated into various mechanical systems, machines traditionally lacking autonomy will become more like robots. This is currently seen for cars with the introduction of semi-autonomous and autonomous vehicles, which include autonomy-focused features beyond what Advanced driver-assistance systems (ADAS) have provided \cite{6936444}. Evidence of this integration can be seen in the ISO 13482:2014 definition for Person Carrier Robot \cite{iso_iso_2011-2}, which shares functionality with autonomous vehicle (i.e. transport a person from one place to another). As autonomous systems with potentially diverse physical bodies start to interact directly with human operators or users, safe operation becomes a critical task \cite{Eder2014}.

Principles from human-robot interaction (HRI) can benefit the design of better interaction with autonomous or semi-autonomous systems. As HRI considers the process in which two agents or more (i.e. robot and human) are interacting with each other and with an environment simultaneously, a parallel can be created for interaction with highly advanced vehicles; in the case of interaction of the driver in a vehicle, the agents share the same physical space (i.e. human inside the vehicle), which limits any interaction with the environment to the vehicle's physical capabilities (e.g. sensing through windscreen viewing angles, actuation through lateral and longitudinal car motion). The human actions are constrained by the cockpit's space and its interfaces (e.g. steering wheel, pedals).

When two agents interact, understanding of the other agent's intentions and decision-making process is essential for seamless interaction \cite{grigore_joint_2013}. This concept often used in HRI is one drawn from the Theory of Mind, which comes from studying human-human interactions \cite{scassellati_theory_2002}. By knowing how an agent is making a decision and what they aim to do, other agents can adapt as to avoid interference (e.g. change motion path to avoid crashing whilst moving) and allow cooperation (e.g. help reach for an object); this information can be drawn from movement patterns. Although variability is present in any task that involves motion \cite{gaudez_intrinsic_2016}, learned tasks such as driving follow known movement patterns that can be expected to be repeatable with some degree of accuracy \cite{schmidt_literature_2014}. Moreover, each person has their distinct movement patterns based on their physical constraints, regardless of variability from movement to movement. These concepts were explored experimentally by recording test subjects whilst driving in a simulated environment.

A range of behaviour identification work related to human motion has been carried out in the form of pose estimation \cite{Kendall_2015_ICCV} (i.e. recognize the human limbs' positions relative to their body) and activity recognition \cite{6208895} (i.e. identify the current task being performed or a description of it). Besides applications focused on computer vision and data analysis problems, motion predictability for HRI has been explored as to include human pose in the robot's decision-making process (e.g. arm motion prediction whilst operating a two-link arm \cite{choudry_stochastic_2013} and anticipatory haptic assistance \cite{medina_synthesizing_2015}, detecting a human reaching for an object \cite{perez-darpino_fast_2015} and whilst moving in an open space \cite{iqbal_movement_2016}); for vehicle and driver analyses, attempts to parametrize driving parameters such as driver behaviour \cite{berndt_continuous_2008-1} and driver intention \cite{oliver_graphical_2000-1} have been achieved without considering an interaction framework (i.e. human driving the actions). HRI approaches consider the human user and the machine together, as they attempt to perform tasks ultimately for the human's benefit. The human is considered the centre of the interaction as it points towards achieving a goal (e.g. health, well-being, labour, entertainment). By creating a more complete analysis that includes the interaction's driving force (i.e. human), richer solutions can be created in contrast to the ones achieved by focusing on the autonomous agent's capabilities and technical developments alone. Basic integration of driver information into a control framework has been proved feasible (e.g. Gray \cite{gray_stochastic_2013} faced sensor noise and computational-related limitations) and further advances on driver intent detection have been employed in the field of autonomous vehicles with rich sensor capabilities \cite{pmlr-v87-casas18a}; nevertheless, to the best of our knowledge there is no work about motion and behaviour prediction for driver manoeuvre prediction with an HRI parallel; recently some ideas about HRI for automated driving have been discussed \cite{biondi2019human}.

A data-driven approach was selected to create the models linking sensor measurements with driver behaviour; this approach allows for flexible model creation, relying on collected data from heterogeneous sources. Among modern data-driven solutions, deep learning is a set of technologies that allows the use of complex layers of successive non-linear processing units to learn multiple levels of representations from large amounts of data \cite{schmidhuber_deep_2015}. In this fast-moving area of research, particular mention goes to the successes for accuracy in image classification \cite{he_delving_2015} and natural language processing tasks \cite{xiong_microsoft_2017}. One type of deep learning algorithm that handles sequences and time-dependent data are recurrent neural networks (RNN); by abstracting the relations in a time sequence of data, past events can be used to inform a current state or event. RNNs have been successfully used in modelling time-dependent phenomena such as human dynamics \cite{fragkiadaki_recurrent_2015} and even anticipation of tasks \cite{jain_brain4cars:_2016}. A recurrent neural network was used to model the driving task as a sequence of time-dependent states from vehicle dynamics and human motion.

The main contributions of this work are twofold: firstly, online analysis and prediction of manoeuvres for integration in a vehicle system was achieved. Expanding our previous work \cite{lopez_pulgarin_drivers_2018}, an RNN-based time-sequence model was introduced that allowed to generalize results from one person to the next so that behaviour could be predicted for unfamiliar participants. Additionally, predictions were achieved without using current knowledge of which manoeuvre is being performed, which was previously not achieved. In addition, this new approach manages to perform both manoeuvre identification and prediction using the same model. Predictions are generated for a set of times of granularity equal to one sampling time, achieving more flexibility than previous results (i.e. one fixed prediction time). The new model allows for manoeuvre identification and prediction for general test subjects, enabling further research for its use in real vehicles for broader ranges of drivers. To the best of our knowledge, we are the first to achieve this with the type of human measurements used here. Secondly, the HRI-based model using human limb position previously introduced to classify driver manoeuvres \cite{lopez_pulgarin_drivers_2017} was extended for driver manoeuvre prediction; feature engineering and an experimental test protocol were used, focused on achieving high performance when dealing with unknown test subjects. The advances reported in our paper set the ground for analysing the driving task from a human-centred perspective, enabling task identification and prediction for further integration as in an HRI scenario.

The remainder of this paper will be divided as follows: Section \ref{sec:methods} introduces all methods used in this paper like the technical details of the experimental design, the structure of individual models, model training and testing methodology and driving modelling; Section \ref{sec:results} shows the results of applying the previously mentioned methods, starting with feature selection and followed by showing how manoeuvre identification and prediction were achieved; and Section \ref{sec:conclusions} presents conclusions and future work.

\section{Methods}
\label{sec:methods}

\subsection{Experimental Modelling}
\label{sec:expmod}

An experiment was designed to obtain data of test subjects using a driving simulator to drive over a virtual track. A generic experimental rig was used like the ones in earlier studies \cite{shia_semiautonomous_2014,lopez_pulgarin_drivers_2018} (i.e. an automotive chair, commercial-grade driver input kit (Logitech G27) and an 21 inches LCD screen) together with a Kinect V2 sensor to record body posture. Carmaker \cite{carmaker_users_2014} was used to simulate both driving scenario and vehicle dynamics (i.e. automatic gearbox vehicle). The tracks followed the British Design Manual for Roads and Bridges for lane widths, curve radius and slope values.

Skeletal tracking information was gathered using the Kinect V2 sensor, which provides estimated depth-based skeletal tracking data with a sampling frequency of 30 Hz or time-step of 33 ms. The sensor was placed on top of the display, 1 metre away from the test chair's default position, with an additional tilt of 15 degrees.

Particular care was given to both skeletal tracking stability and test subjects getting accustomed to using the simulator (i.e. conditioning). A sensor calibration protocol was used to ensure skeletal tracking of both arms whilst participants were holding the steering wheel; the test subject stood up next to the chair and then sat down once initial tracking was achieved; this was done due to sensor restrictions (i.e. viewing angles generated ambiguous pose errors if the tracking did not begin whilst standing up). As a way to match all the test subjects' previous driving experience to the simulated environment, a complete track of all possible manoeuvres (i.e. left and right turn, drive straight) was performed before the test. Before and after the driving test a Lane Change Task (LCT) of 1800 m \cite{petzoldt_learning_2014-1} was implemented to confirm driver conditioning. LCT is a standard test used in driver distraction testing, which consists of successive lane changes signalled by traffic signs at the side of the road; the driver must perform a lane change manoeuvre every 150 m whilst keeping a cruise speed of 60 km/h, with the sign being visible only 40 m before the car reaches the traffic sign; a mean deviation (MDEV) measured from the driver's position against a normative lane change model served to measure the driver's conditioning. An MDEV $< 0.7$ m was achieved in all test subjects, as recommended by the technical norm ISO (ISO/DIS 26022, 2010). In general, each test drive consisted of approximately 20 minutes, for which data were recorded at the previously mentioned sampling frequency.

Carmaker provided information about the parametrized vehicle and its surrounding environment. Virtual sensors generated a complete description of the vehicle state and its surroundings, such as vehicle dynamics (e.g. velocity, acceleration, engine revolutions, orientation) and driver input (e.g. steering wheel position, pedals' input). Scenario creation was performed by parametrically describing sections of the road (e.g. length, slope, turn angle), which in turn created a parametric map of the testing road.

\begin{figure}[ht]
	\centering
	\includegraphics[clip,width=0.9\columnwidth]{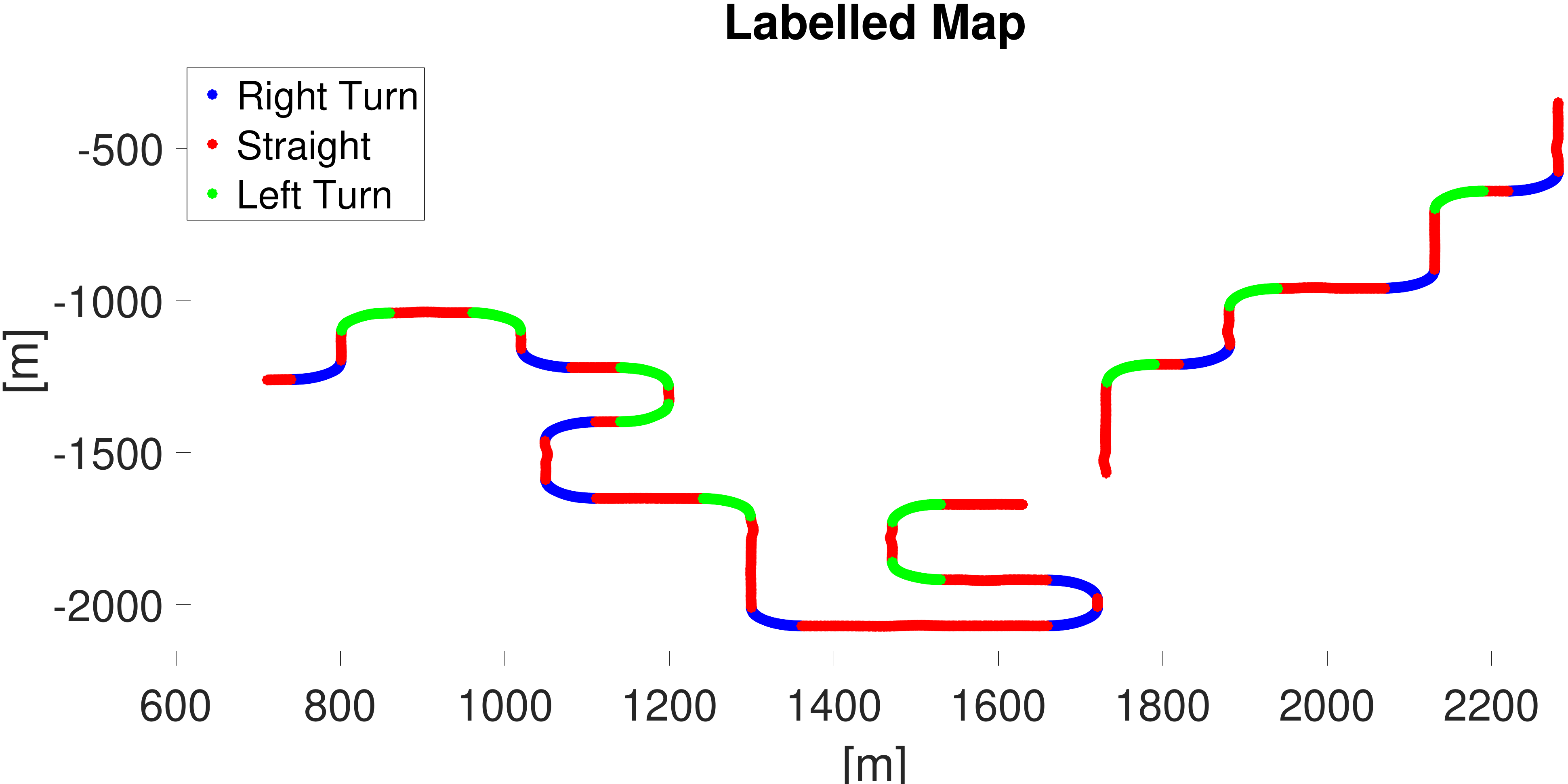}
	\caption{Labelled road map}
	\label{fig:map}
\end{figure}

Building on the idea of creating a real-time system to analyse data online, a labelling algorithm was implemented to label each data point gathered during the test. By using Carmaker's inner scenario description, all road sections were labelled based on the parametric map describing the sequential set of road segments (e.g. straight 500 m, left turn radius 30 m); the virtual position of the car was then compared with the virtual map, to assign a label to the driver's data as seen in Figure \ref{fig:map}. This approach allowed the creation of precisely labelled datasets, reducing the variability of results that could be introduced by labelling errors and setting a clear threshold for manoeuvre transition. By this means, the set of labels \(y \in M = \{M_1,\dotsc,M_n\} \) was created, with $M_{n}=\{0,1\}$ as the set of classifiable manoeuvres consisting of $M_1$ as left turn, $M_2$ as straight manoeuvre and $M_3$ as right turn.

Twenty-nine test subjects (17 females, 12 males) took part in the experiment. None of them had previous experience with driving simulators but did have good enough real driving experience (mean=9.36 years, min=3, max=39, standard deviation (SD)$=7.44$). All participants confirmed to have normal or corrected-to-normal vision and a valid driver’s licence.

Test subjects' physical dimensions and chair preferences were taken, as it validates how generalizable the results are based on the test subjects' variability; arm length (mean=54.88 cm, min=44, max=62, SD=3.88), torso length (mean=43.24 cm, min=38, max=53, SD=3.83) and the relative distance between chair and the screen (mean=5.52 cm, min=-1, max=17, SD=5.77) were taken. Experiments were approved by the Ethics Committee of the Faculty of Science, University of Bristol ID 20168941282. Participants gave their informed written consent prior to participation. 

\subsection{Model Structures}
\label{sec:ms}
Data-driven techniques in the form of machine learning algorithms were implemented to create models that allow for driving manoeuvre identification (i.e. current manoeuvre) and prediction (i.e. future manoeuvre). Machine learning has proven useful in HRI and vehicle related tasks in its different approaches (e.g. supervised learning with both regression \cite{crabbe_skeleton-free_2015}, classification \cite{berndt_continuous_2008} and reinforcement learning \cite{kumar_learning-based_2013}). Supervised learning was used throughout this work, for the case of mapping input features to output labels (i.e. classification task). Focus was put on the use of recurrent neural networks, but three additional algorithms were included for comparison with previous work \cite{lopez_pulgarin_drivers_2018}.

Recurrent Neural Networks (RNN) retain information about past inputs in a sequence of data; by using a hidden state per time step $h_t$, information from several time steps can be retained. Long Short-Term Memory (LSTM) \cite{hochreiter_long_1997} tackles problems that traditional RNN had (e.g. vanishing/exploding gradient and state retention over long periods); it adds gating units to each memory cell, controlling how each cell state $C_t$ and hidden state $h_t$ are updated in time. An extension of LSTM is the Bidirectional LSTM (BI-LSTM) \cite{schuster_bidirectional_1997}. It exploits past and future information during training to get a better understanding of the data. Hidden layer outputs in both forward and backward structures are computed together and used as the cell output, with cell output $h_{t}$ as $h_{t} = h^a_{t} \oplus h^b_{t}$ and $ \oplus $ as the concatenation operator.

Support vector machines (SVM) find an optimal hyperplane or several hyperplanes that maximize the separation margin between classes (i.e. distance to the nearest training data point of any class) in any dimensional space. Weighted SVM or SVM-W \cite{huang_weighted_2005} introduce a weighting factor that allows for hyperplanes to adjust better to unbalanced classes \cite{lopez_pulgarin_data-driven_2016}.

An extra trees classifier (ET) \cite{geurts_extremely_2006} fits several randomized decision trees on sub-samples of training data; it learns simple decision rules in recursively partitioned data and fits similar rules to each subset \cite{berk_classification_2016}, thus obtaining better performance compared to traditional decision trees.

Multi-layer perceptrons (MLP) \cite{he_delving_2015} are classic neural networks where information flows from the input layer to output layer, learning a non-linear approximation between the former ones.

\begin{figure}[ht]
	\centering
	\includegraphics[clip,width=0.98\columnwidth]{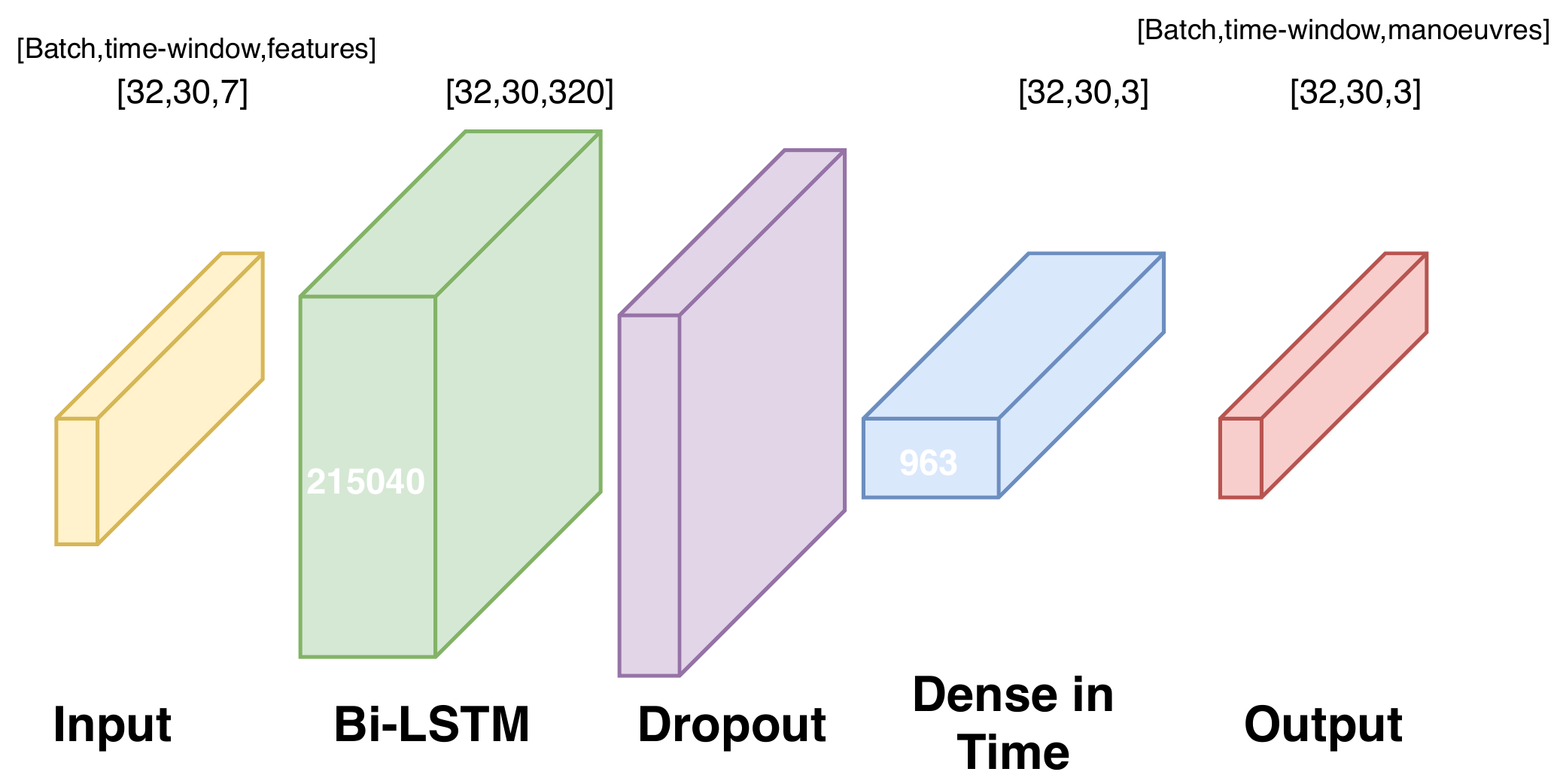}
	\caption{Proposed structure for BI-LSTM model with its layers, layer shapes (i.e. training batch, prediction time window and dimension input/output)}
	\label{fig:lstm_s}
\end{figure}

The parameters used for the models were: for BI-LSTM, the structure is shown in Figure \ref{fig:lstm_s} for a “many-to-many” prediction, with 160 LSTM units returning sequences, dropout of 0.2, softmax function activation for the time layer, categorical cross-entropy loss function, adam optimizer \cite{kingma_adam:_2014} and 216003 parameters in total; for SVM, a radial basis kernel, multi-class strategy of one-vs-one and class weighting were used with class weights inversely proportional to the frequency of a class; for ET, we used 10 estimators and 2 leafs; for MLP, we used rectified linear unit (ReLu) activation functions and cross-entropy loss function. All parameters were determined through experimentation. Selection of these parameters came from experimentation within different ranges (e.g. BI-LSTM with units $\{60,120,160,240\}$, state and stateless, LSTM and BI-LSTM, input sequence length $\{15,30,45\}$) and good practices for recurrent neural networks (e.g. dropout layer).

\subsection{Model Training and Testing Methodology}
\label{sec:mttm}

The proposed tests focused on validating the model's ability to produce accurate results for both known and unknown test subjects, as data can be very different even if drivers were performing the same task. This is the reason why manipulation of both amount and source (i.e. which test subject) of training and test data was vital for success. For this purpose, the following tests were designed considering both a model $S$, specified by the test subjects' data used for its creation, and the validation set $ts^{V}$ (i.e. data for which it is tested against). This means using a model $S_{i}$ and test set $ts^{V}_{i}$ for any single individual or a model $S^{C}_{a,\dotsc,b}$ and test set $ts^{V}_{a,\dotsc,b}$ for any set of individuals with $i=\{1,2,\dotsc,N\}$ and $N=29$.

\begin{itemize}
	\item \textbf{Test 1:} Evaluates an individual model's feasibility and generalization capability to known test subjects (i.e. itself). $S_{i}$ is tested against $ts_{i}^{V}$ for the same $i$.
	\item \textbf{Test 2:} Evaluates an individual model's generalization capability to unknown test subjects. $S_{i}$ is tested against $ts_{j}^{V}$ for all subjects $j \neq i$ in $j=\{1,2,\dotsc,N\}$.
	\item \textbf{Test 3:} Evaluates a concatenated model's feasibility and generalization capability to known test subjects (i.e. itself). $S^{C}_{a,\dotsc,b}$ is tested against $ts_{a,\dotsc,b}^{V}$ with the same combination of subjects $a$ and $b$ for $a=1$ and $b=\{1,2,\dotsc,N\}$.
	\item \textbf{Test 4:} Evaluates a concatenated model's generalization capability to unknown test subjects. $S^{C}_{a,\dotsc,b}$ is tested against $ts_{c,\dotsc,d}^{V}$ for $a=1$, $b=\{1,2,\dotsc,N\}$, $c=b$ and $d=\{c,c+1,\dotsc,N\}$.
\end{itemize}

Stratified (i.e. separated by classes) training and validation sets were created for randomized cross-validation tests, with 70$\%$ training data and 30$\%$ test data. F1 score metric was used to measure model performance as a high F1 score equates to both high precision and high recall \cite{sokolova_systematic_2009}.

\subsection{Driver Modelling}

\subsubsection{Manoeuvre Transition}

\begin{figure}[ht]
	\centering
	\includegraphics[clip,width=0.75\columnwidth]{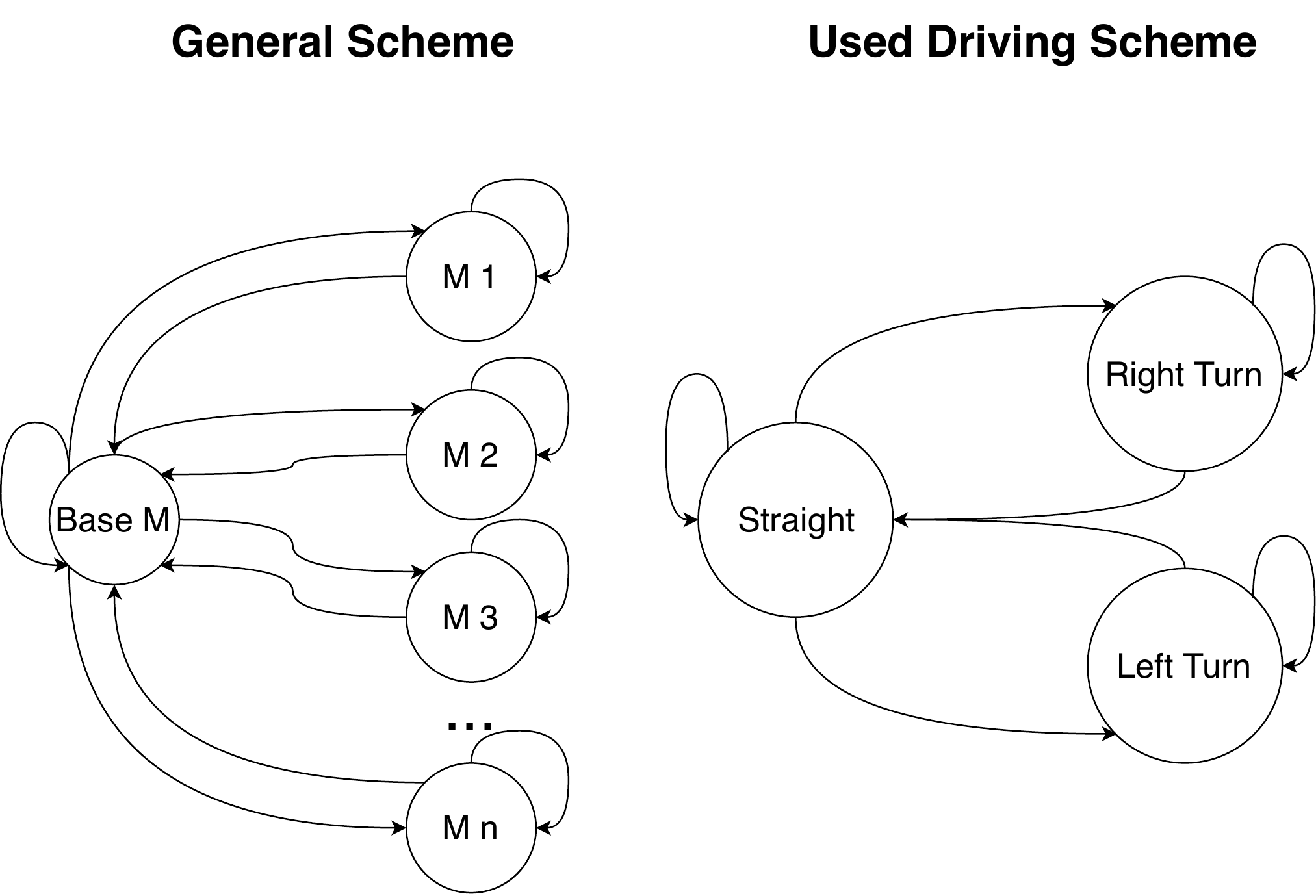}
	\caption{State transition for driving manoeuvre modelling}
	\label{fig:manoeuvre_sts}
\end{figure}

Manoeuvre-oriented driver modelling is a way to reduce the complex problem of human driving and general car navigation into tasks or states that relate to each other. The idea behind this process comes from the study of movement and behaviour in fields such as neuroscience; by taking a complex process with large sensory inputs and reducing it to a set of states of the system \cite{wolpert2000computational}, computationally feasible models can be generated with results reflecting the system itself. A finite-state transition model was proposed to describe the driving manoeuvres, as used in previous work \cite{lopez_pulgarin_drivers_2017, lopez_pulgarin_drivers_2018}. Figure \ref{fig:manoeuvre_sts} shows a general scheme, expandable to $n$ number of complex manoeuvres branching from a base manoeuvre; a reduced set was defined, with going straight as the base manoeuvre, and subsequent turn manoeuvres as other possible states (i.e. left turn and right turn) with equal state transition probability. Hence, by selecting the three state model, we are following the sampling based / statistical learning concept of perception in neuropsychology \cite{lengyel2019unimodal}; it argues that simple, even uni-modular perception and decision models are the basis of representations and further generalizations of complex reality.

Based on that assumption, it can be argued that general driving can be restricted to a defined set of connected manoeuvres. Each manoeuvre can be succeeded by the same or a different one. This manoeuvre modelling scheme works well with the labelling mechanism explained in Section \ref{sec:expmod} and was the basis for using manoeuvres as labels in a machine learning classification task; label binarization was applied to manoeuvre labels $y$ creating $ \tilde{y} =\begin{bsmallmatrix} 1 & \cdots & 0\\ \vdots & \ddots & \vdots \\ 0 & \cdots & 1 \end{bsmallmatrix} $ for any $n$ manoeuvres $M_n$.

\begin{figure}[ht]
	\centering
	\subfloat[Frontal View]{%
		\includegraphics[clip,width=0.31\columnwidth]{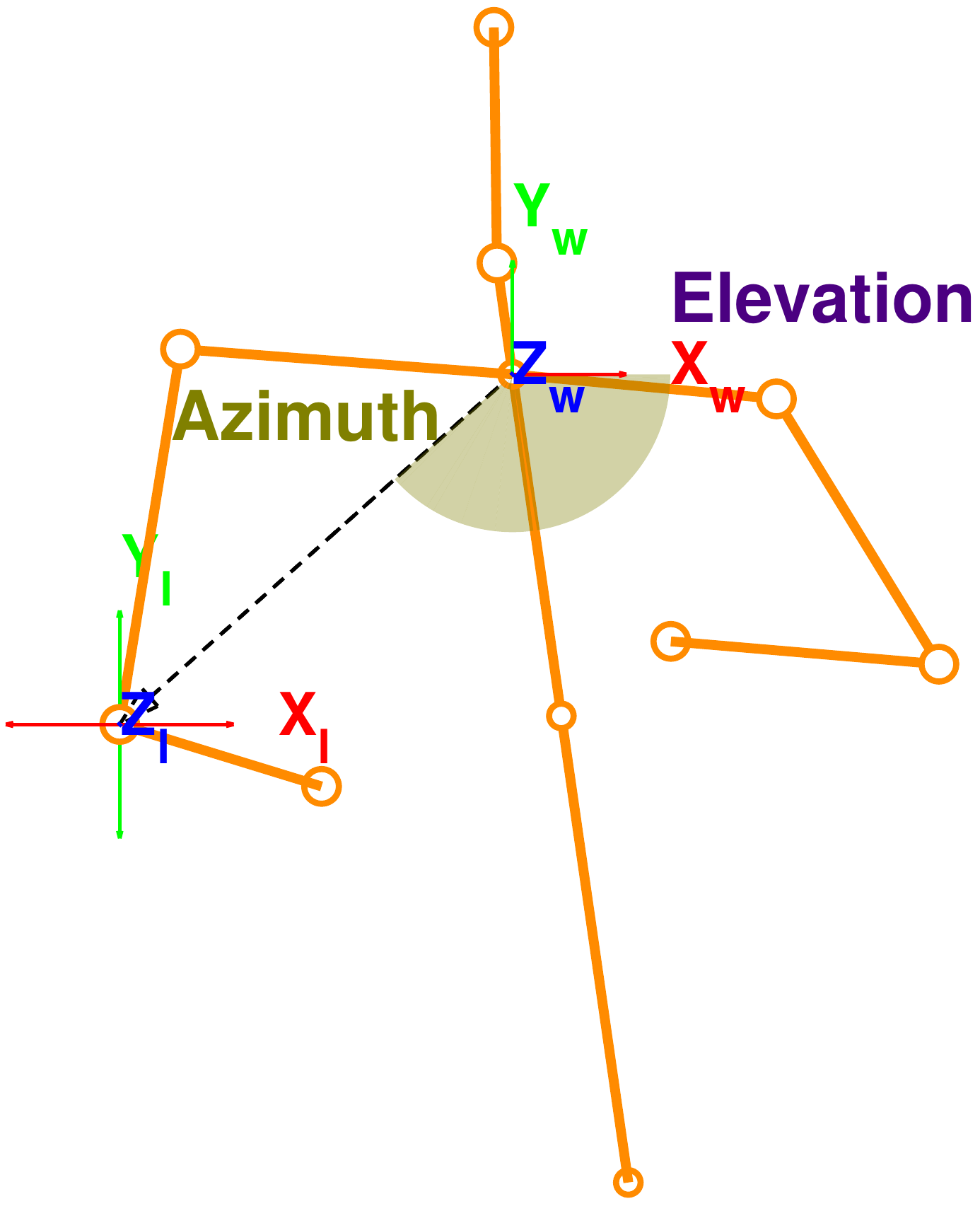}
	}
	\subfloat[Upper View]{%
		\includegraphics[clip,width=0.31\columnwidth]{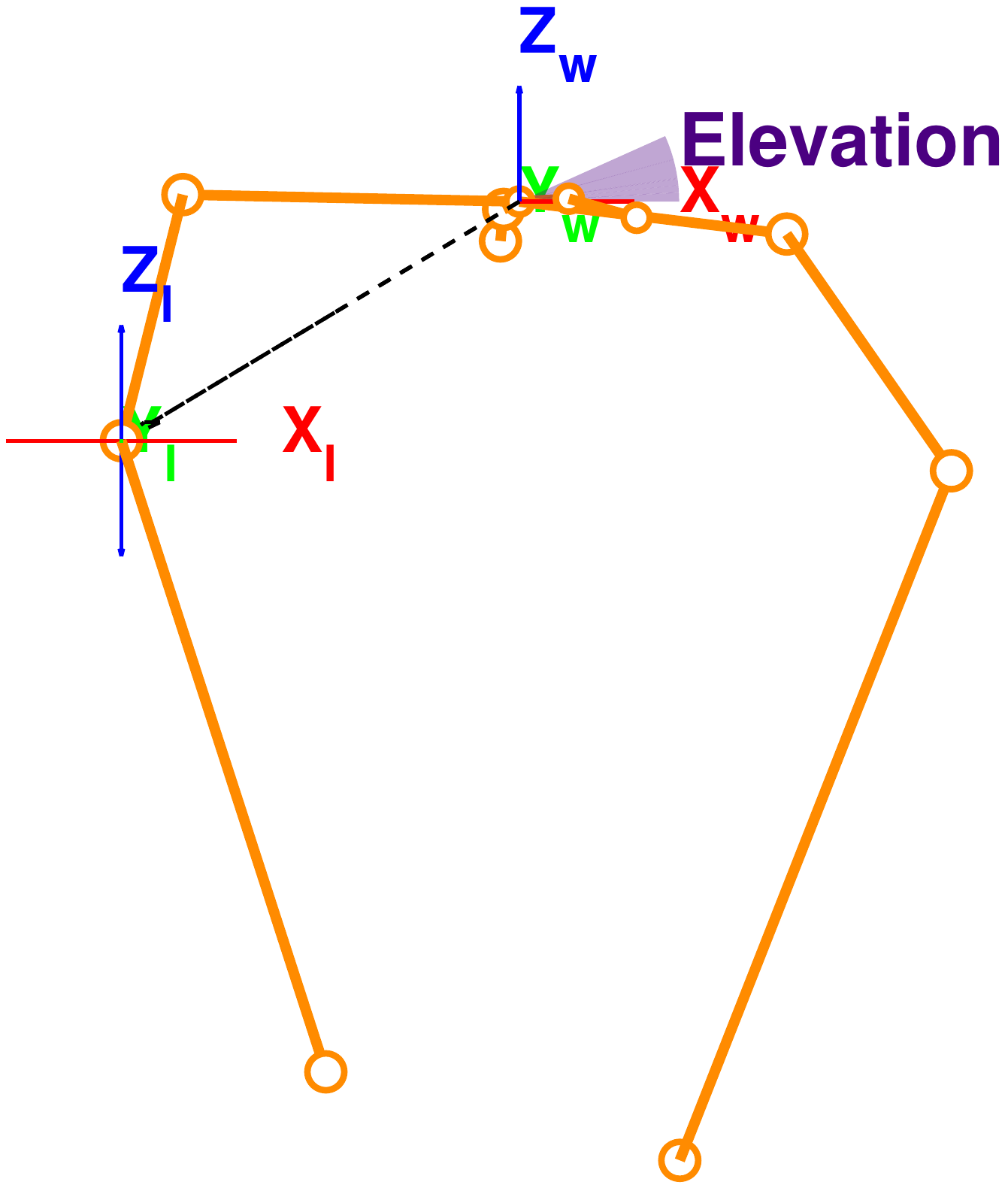}
	}
	\subfloat[Side View]{%
	\includegraphics[clip,width=0.33\columnwidth]{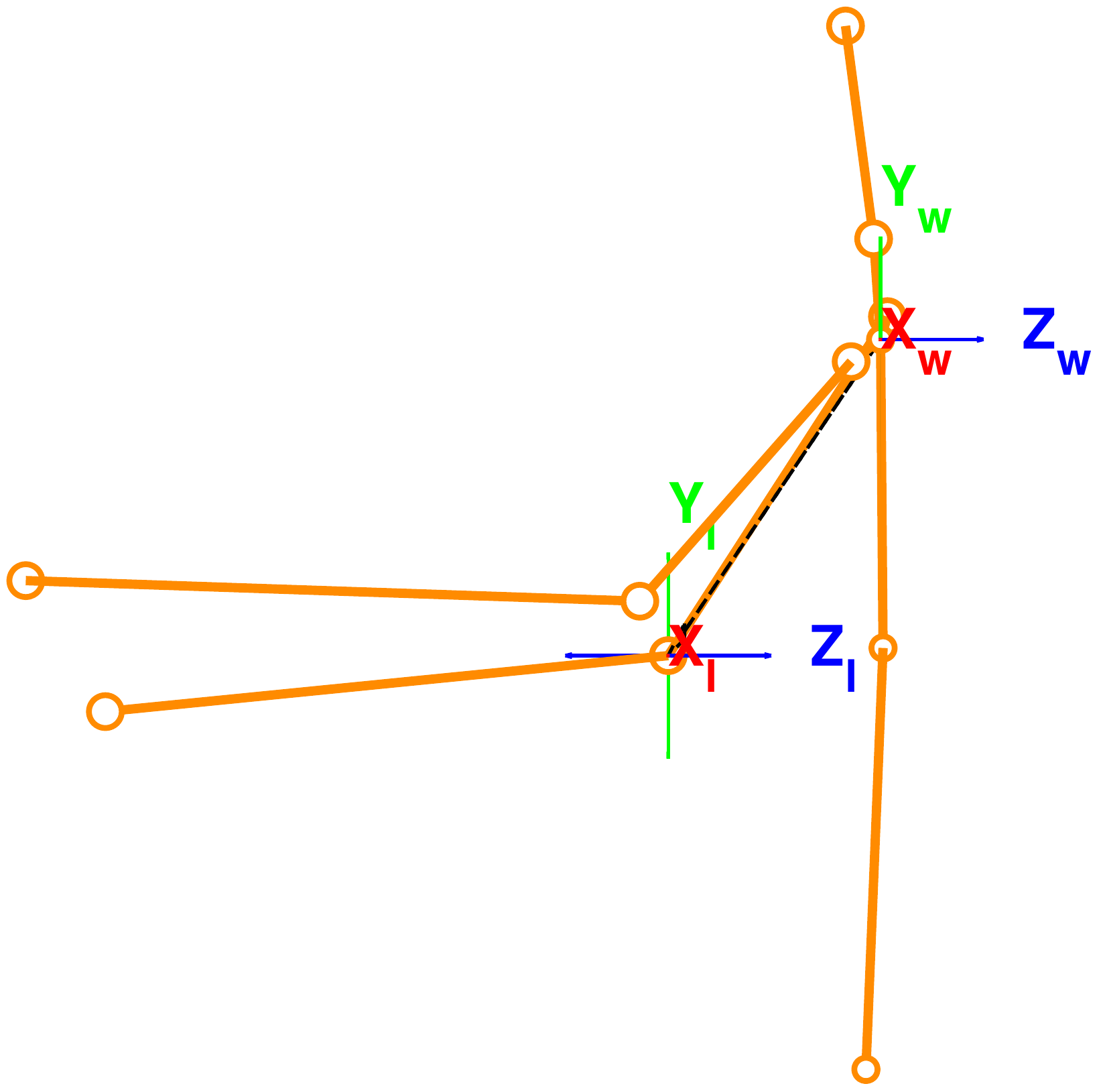}
	}
	\caption{Spherical projection of the angle between elbow and shoulder}
	
	\label{fig:img1}
\end{figure}

\subsubsection{Skeletal Tracking}

Skeletal tracking systems estimate information about human pose in terms of the position of limbs relative to the body and the sensor; for this work, we used the Kinect V2 RGB-D camera with the standard skeletal tracking proprietary software. Kinect V2 provides the position and orientation of 25 joints in total, based on a modified algorithm that uses depth measurements and starts calculations from the base of the identified torso continuing to the end of each limb \cite{shotton_real-time_2011}). Being a consumer-grade sensor designed for gaming (i.e. depth range and field of view suitable for open rooms), the use of Kinect V2 at short distances with cluttered environments raised problems such as occlusion and out-of-range measurements.

However, based on claims from the field of ergonomics and bio-mechanics that both shoulder and elbow position contain crucial information for driving, we focused on measurements above the hands only. These positions have been proved to influence steering speed and accuracy \cite{schmidt_influence_2015} with a high muscle activation rate in those regions alone whilst driving \cite{liu_function_2012}.

Estimated elbow position presented high levels of noise and jitter in the z-axis, due to the previously mentioned problem of ambiguous viewing angles. This noise came from the estimation process done by the sensor, which estimates any possible value for a joint's position, even if generating outliers and signal discontinuities. To tackle this challenge, a spherical projection of the angle between the elbow and shoulder with relation to the midpoint between the shoulders was proposed; only the two angular components of the projection were considered, as the radial component was more related to depth changes and consequently to the noise previously described. Figure \ref{fig:img1} shows these angles in skeletal tracking data of a test subject whilst driving, with both base reference frame ($X_w,Y_w,Z_w$) and the elbow reference frame ($X_l,Y_l,Z_l$).

The above method was selected due to the nature of the steering task which can be seamlessly described by angular movements. This reduced the overall variance of the signal by $42\%$ and standard deviation by $62\%$ compared to the raw measurement.

The resulting feature vector was further processed to improve the performance of some models. For BI-LSTM, it was scaled to a range between $(-1,1)$. For SVM, it was scaled to a unit range, with variance and mean removed (i.e. standardization). Both processes were learned from the training set only, and are applied in all remaining tests.

\section{Results}
\label{sec:results}

\subsection{Feature Selection}

Due to the temporal nature of the measurements, selection of a feature set $z^{c} \in \chi = \mathbb{R}^{n}$ that encodes enough information about all acting agents is crucial.

Available data came from three heterogeneous data sources: skeletal tracking data from the test subject (i.e. Human-based data), steering wheel and pedal measurements (i.e. device which allow Human Car Interaction or Interaction-based data) and virtual measurements from the simulated car and its sensors (i.e. Car-based data).

Desired features should allow the models to obtain high accuracy for all known test subjects whilst retaining the ability to generalize to unknown test subjects (i.e. no over-fitting). As feature selection is a non-trivial process highly dependent on the data available, a set of test features were handcrafted and validated experimentally. The following sets were designed and tested:

\begin{itemize}
	\item \textbf{SET1:} Upper torso / limb position.
	
	\begin{itemize}
		\item \textbf{Spine Shoulder:} Relative position of the spine at the height of the shoulders measured from the sensor's position (X, Y, Z).
		\item \textbf{Left and Right Shoulder} Relative position of the left and right shoulder joint measured from the sensor's position (X, Y, Z).
		\item \textbf{Left and Right Elbow:} Relative position of the left and right elbow joint measured from the sensor's position (X,Y,Z).
	\end{itemize}
	
	\item \textbf{SET2:} SET 1 minus spine shoulder position.
	\item \textbf{SET3:} Skeletal features.
	
	\begin{itemize}
		\item Azimuth component of the spherical projection for the angle between elbow and shoulder with relation to the base of the torso per arm.
		\item Elevation component of the spherical projection for the angle between elbow and shoulder with relation to the base of the torso per arm.
	\end{itemize}

	\item \textbf{SET4:} SET 3 adding the driver input and robot measurements.
	
	\begin{itemize}
		\item \textbf{Steering wheel angle}.
		\item \textbf{Gas pedal level}.
		\item \textbf{Car Velocity:} Current speed of the vehicle.
	\end{itemize}
	
	\item \textbf{SET5:} SET 2 adding the driver input and robot measurements of SET 4.
	
\end{itemize}

\begin{figure}[ht]
	\centering
	\includegraphics[clip,width=0.98\columnwidth]{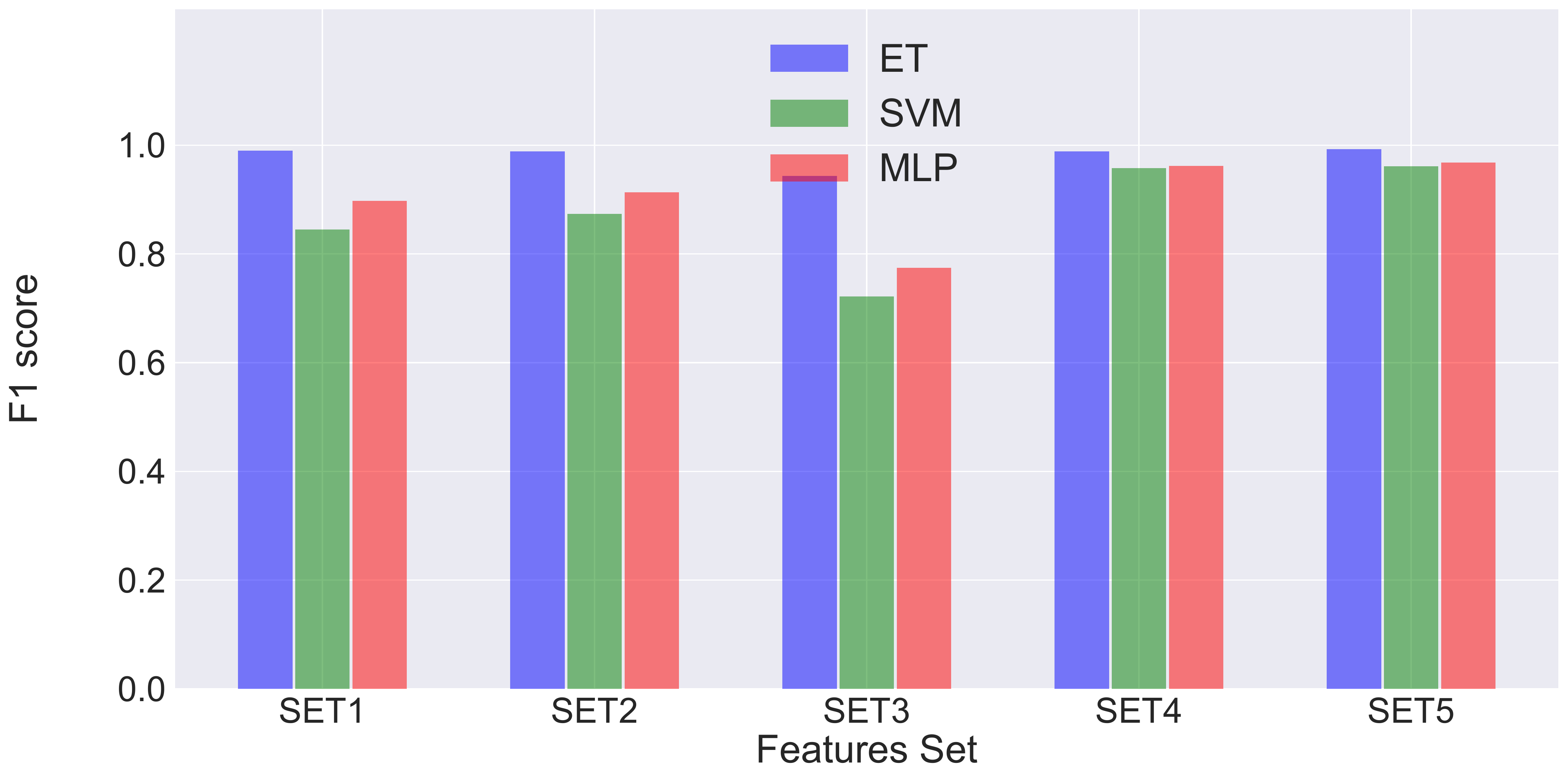}
	\caption{Feature test 3 results averaged across all test subjects}
	\label{fig:img2a}
\end{figure}

\begin{figure}[ht]
	\centering
	\includegraphics[clip,width=0.98\columnwidth]{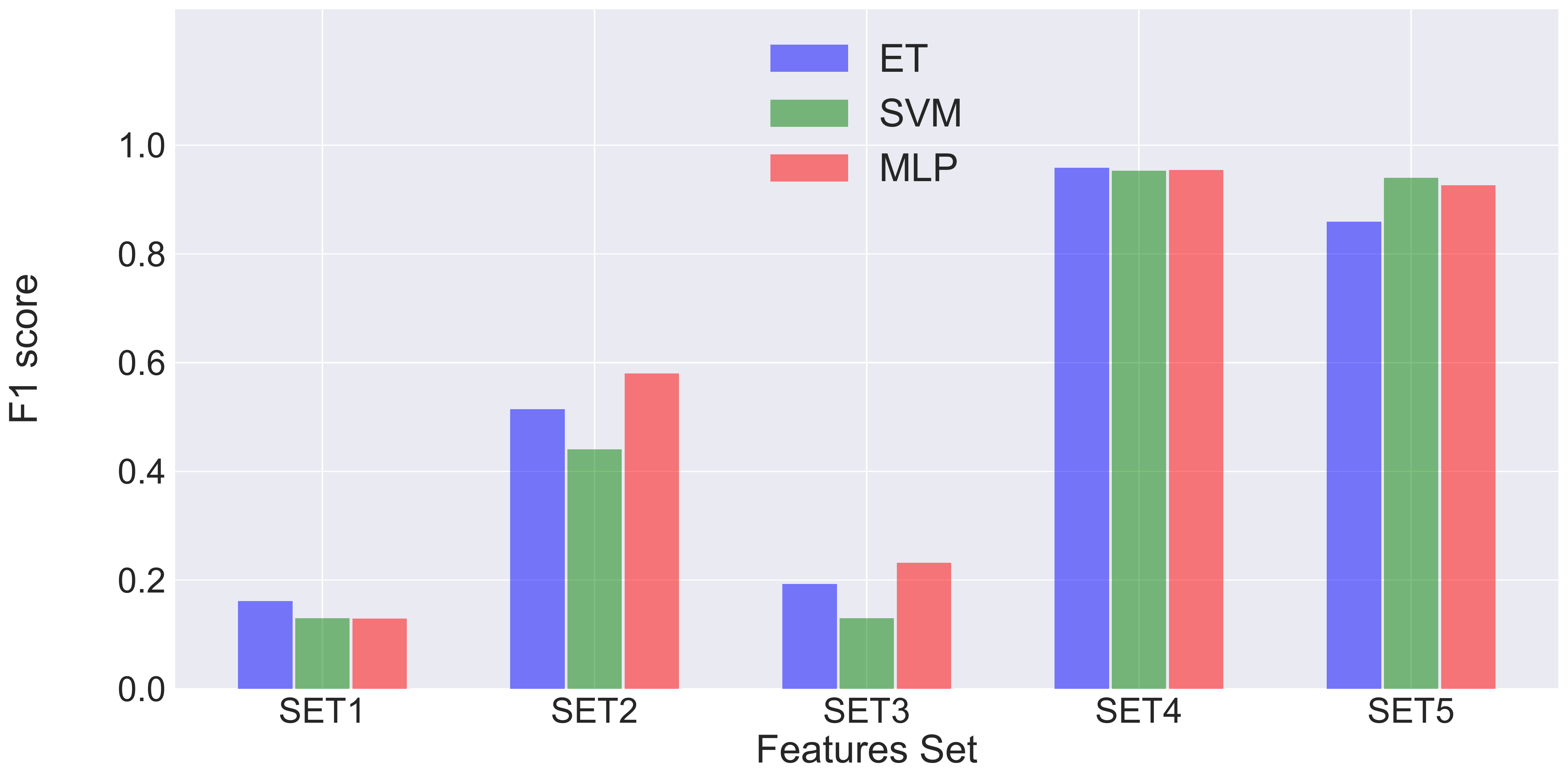}
	\caption{Feature test 4 results averaged across all test subjects}
	\label{fig:img2b}
\end{figure}

These sets were used to train shallow learning models for a manoeuvre identification task and tested based on the procedure explained in Section \ref{sec:mttm}; focus was given on testing how adaptable and flexible a resulting model could be to different known and unknown test subjects, looking to translate these capabilities to manoeuvre prediction. Data from 25 test subjects were used for training/test validation of known test subjects, and the remaining 4 test subjects were used for unknown test subject validation (i.e. Test 3 and Test 4 mentioned in Section \ref{sec:mttm}). Figure \ref{fig:img2a} (Test 3) shows results above $75\%$ for all combinations, with better performance for those in which the heterogeneous signals were included (i.e. SET4 and 5). For both tests, SET2 (i.e. centred joint coordinates) performed better than SET1 (i.e. raw joint positions). SET3 (i.e. skeletal-based features) did not seem to perform better in either of the tests when compared to SET1 and SET2, contradicting the initial idea of better performance by reducing signal noise. However, Test 4 in Figure \ref{fig:img2b} shows that adding driver input and robot measurements to skeletal-based features (i.e. SET4) managed to generalize better across subjects than any other combination of data. This result confirmed the need to explore the input features to achieve a good result when using heterogeneous data sources.

After validating the input feature vector, it was used to create and validate models for identification and prediction of manoeuvres for known and unknown test subjects.

\subsection{Manoeuvre Modelling}
\label{sec:mm}

The BI-LSTM model was used with the parameters described in Section \ref{sec:ms}. Models were trained and tested following the methodology introduced in Section \ref{sec:mttm}. Model input came as a time-sequence of features or past data, with a length of $t_{wi}=30$ time steps; an individual time step is equal to 33 ms. The model's output is a time-sequence of probabilities, referring to how likely it is that each type of manoeuvre is or will be performed; the output considers a prediction window or $t_{wo}$ of 30 time steps after present time; output at time step $t_{wo}=0$ can be interpreted as the current manoeuvre being performed and time step $t_{wo}>0$ as the probability of a manoeuvre to be performed a certain amount of time-steps after the current manoeuvre.

\subsubsection{Test 1}
\label{sec:mm_1}
Predictions for individual subjects were tested as explained in Section \ref{sec:mttm}. Individual models (i.e. models trained with data from only one subject $S_{i}$) tested against a known test subject $ts_{i}^{V}$. Each model was able to anticipate the coming manoeuvre several time steps before the actual manoeuvre is performed, with predictions as much as $0.9$ seconds before it was performed. Analysing performance by manoeuvres showed results higher than $80\%$ for most test subjects and manoeuvre combinations; note that this was achieved despite the relatively small training set per test subject available (e.g. each test drive was not longer than 20 minutes). There are differences in manoeuvre-based performance between test subjects (e.g. left and straight manoeuvre tendency to score higher than right manoeuvre for some test subjects), but there was no significant difference for the majority of participants that would justify a manoeuvre-based analysis; reasons behind this variability could come from sensing errors from the skeletal tracking information, as no other data-source (e.g. car dynamics) behaves differently based on the manoeuvre performed. Based on these results, further reports will show performance averaged across the three types of manoeuvres used in this investigation.

\begin{figure}[!ht]
	\centering
	\includegraphics[clip,width=\linewidth]{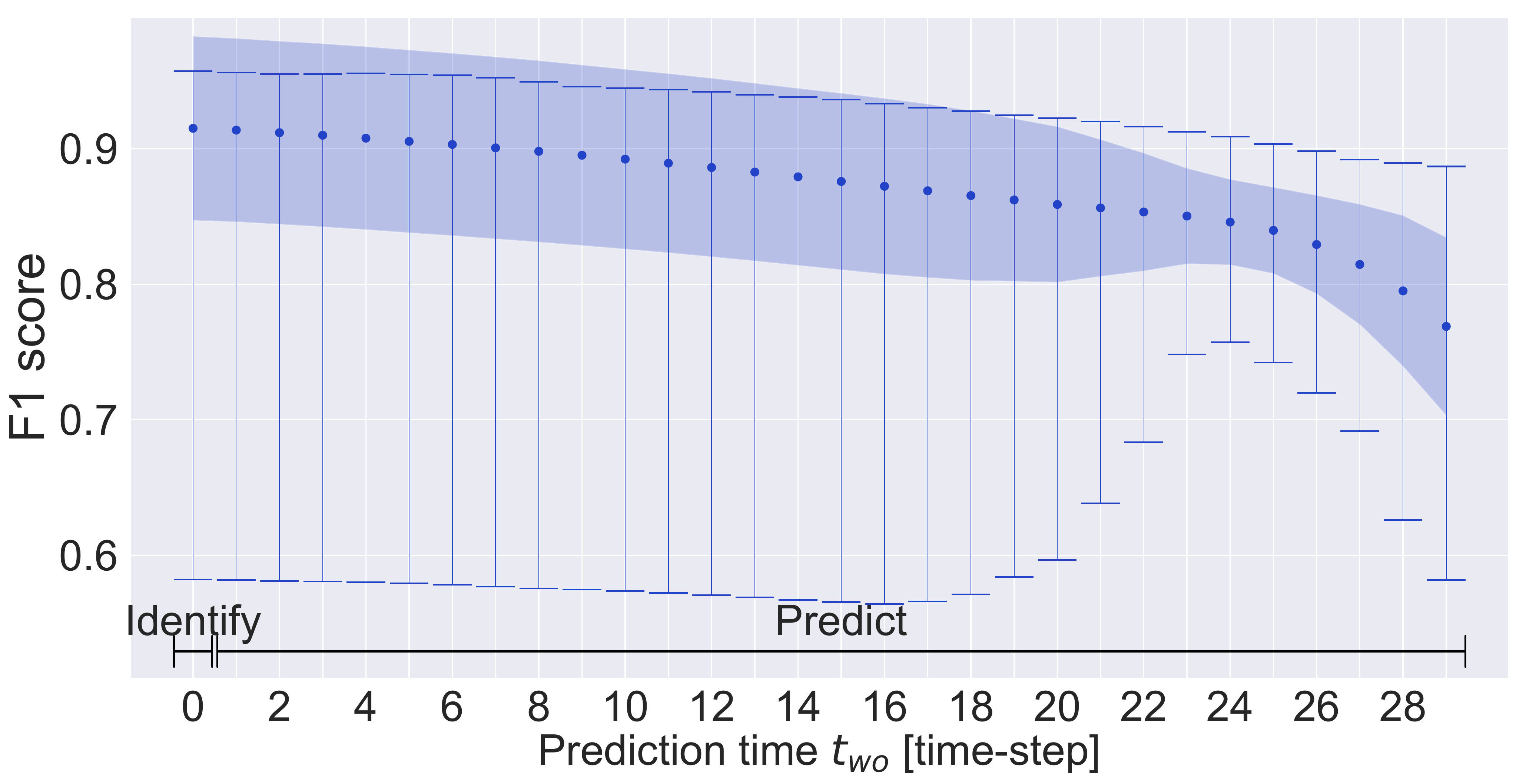}
	\caption{Statistical analysis of F1 metric for models $S_{i}$ tested against test subjects $ts_{i}^{V}$ for $i={1,2,\dotsc,N}$. Average of all $i$ for each prediction time (dot), standard deviation (shaded area), max and min values (caps)}
	\label{fig:pred5}
\end{figure}

Considering that the model output is a temporal sequence with different outputs per time-step, each of them relates to a distinct prediction in time. It was expected that manoeuvres more distant in the future were harder to predict, but the specifics of how these work for each test subject were not intuitive. Hence, a statistical analysis was performed to confirm the performance trends per time step inside the analysed time window. Figure \ref{fig:pred5} shows prediction performance per $t_{wo}$ of all models tested $S_{i}$ with data from their respective test subjects $ts_{i}^{V}$; performance was measured using F1 metrics averaged across manoeuvres and among all test subjects; maximum, minimum and standard deviation of the averaged results are displayed. Figure \ref{fig:pred5} confirms both that performance decreases as $t_{wo}$ increases and that high data variability exists among test subjects. Results' dispersion is high for most $t_{wo}$, as seen by the standard deviation being larger than the maximum values (e.g. for $t_{wo}=[1,\dotsc,18]$) and by the big difference between the mean value and its extremes (e.g. for $t_{wo}=[1,\dotsc,22]$); these results highlight the performance differences among individual models $S_{i}$, which can be attributed to their training dataset and each individual's driving patterns (i.e. behaviour before a certain manoeuvre differs between test subjects and within test subjects for the same manoeuvre).

\begin{figure}[!ht]
	\centering
	\includegraphics[clip,width=\linewidth]{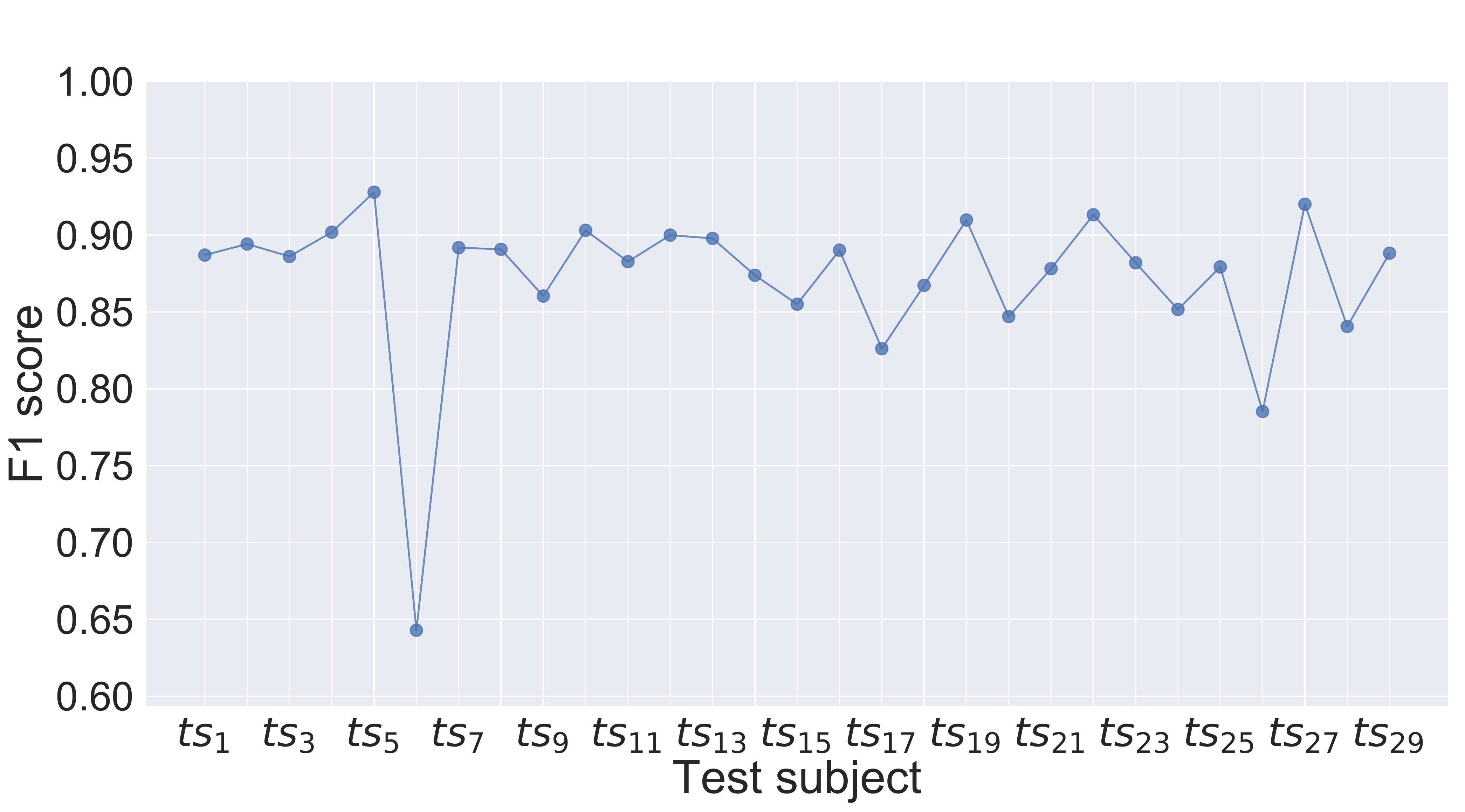}
	\caption{Mean F1 metric for models $S_{i}$ tested against test subjects $ts_{i}^{V}$ for $i=\{1,2,\dotsc,N\}$. Results were averaged over all prediction times and manoeuvres}
	\label{fig:pred6}
\end{figure}

Finally, the overall performance of the models was assessed. Figure \ref{fig:pred6} shows prediction performance of all individual models $S_{i}$ against $ts_{i}^{V}$; performance was measured using F1 metrics averaged by prediction window and manoeuvre. This graph confirms the individual model’s ability to model manoeuvres from one test subject for the presented time steps' with a performance above $80\%$ in most cases.

Note, however, that despite most models managing to model manoeuvres with small amounts of data from one specific test subject (e.g. $S_{2}$), performance among these models differed greatly for $t_{wo}=29$ (e.g. $S_{2}$ vs $S_{28}$) and some others greatly underperformed when compared to the majority (e.g. $S_{6}$, $S_{26}$) as seen in Figure \ref{fig:pred6}. Both findings can be connected to intra-individual variability whilst driving (i.e. driving styles not repeatable enough as to get enough information to understand a certain test subject's driving style), sensor noise from skeletal tracking data and the BI-LSTM's inability to create meaningful predictions with small amounts of data \cite{schmidhuber_deep_2015}; both situations can be tackled by adding more data to the training set from the test subject in question.

As manoeuvre identification and prediction for known test subjects for a single model was confirmed as feasible, further analyses of how this model can generalize to data from unknown test subjects were performed.

\subsubsection{Test 2}
\label{sec:mm_2}
Predictability of unknown test subjects was tested as explained in Section \ref{sec:mttm}. Individual models $S_{i}$ generalizing to predict behaviour of unknown test subjects $ts_{a}^{V}$ was tested. Most individual models were not able to cope with data of unknown test subjects (i.e. some models performed below chance; e.g. $S_{1}$, $S_{18}$, $S_{26}$). However, some models managed to generalize to other test subjects with results similar to the ones seen in Figure \ref{fig:pred6} (i.e. around $85\%$), which was not achieved using shallow models (i.e. manoeuvre prediction for unknown test subjects).

These results further highlight intra-individual differences in the way participants perform a certain manoeuvre and their behaviour before that particular manoeuvre; these could be attributed to physiological differences between participants (i.e. arm and torso length which influences movement patterns), dynamic aspects related to individual driving style and variance within a participant. Despite such inter- and intra-individual variability, however, our findings indicate that generalization to other test subjects is possible.

\begin{figure}[!ht]
	\centering

	\includegraphics[clip,width=\linewidth]{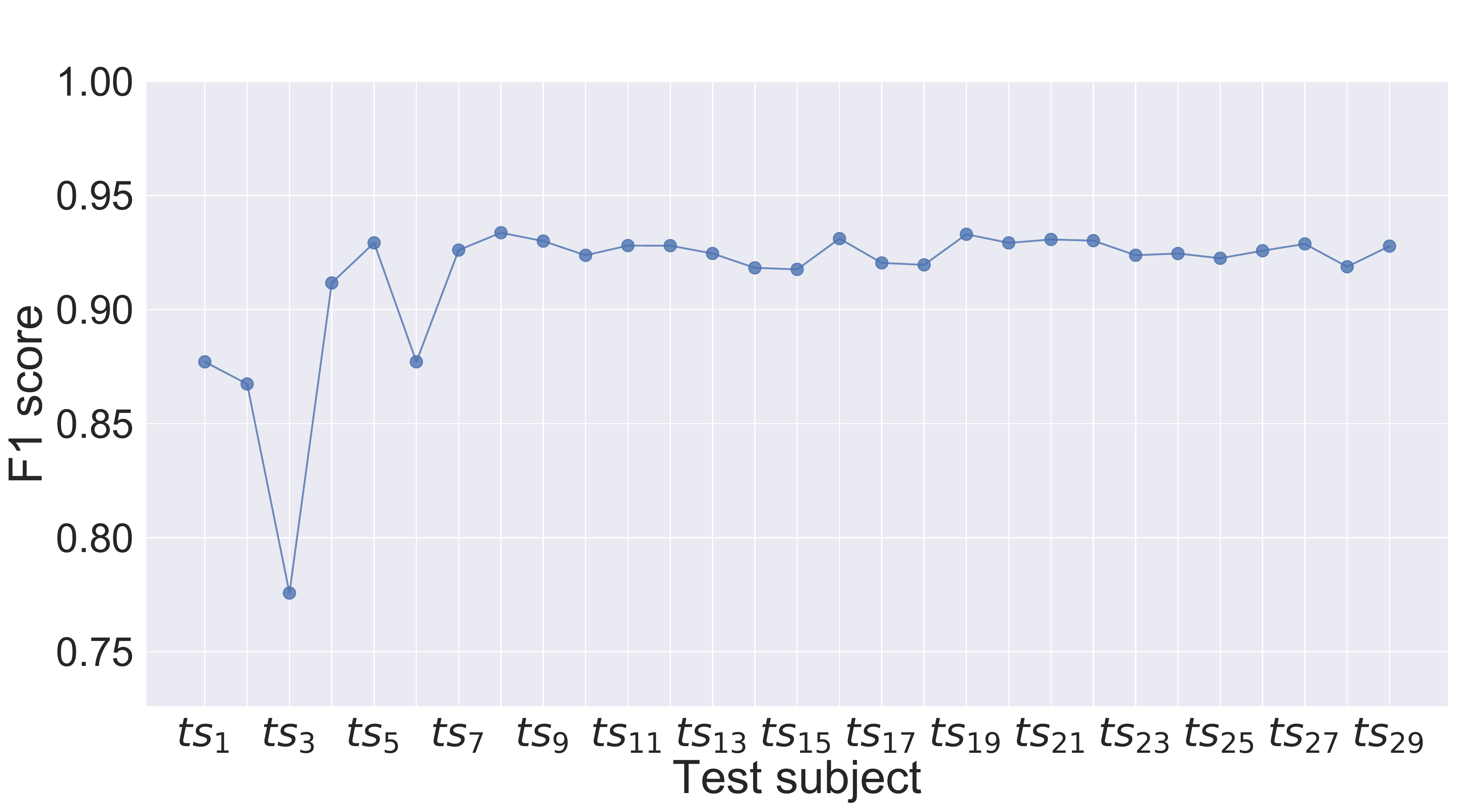}
	\caption{Mean F1 metric for models $S^{C}_{a,\dotsc,b}$ tested against test subjects $ts_{a,\dotsc,b}^{V}$ for $a=1$ and $b=\{1,2,\dotsc,N\}$. Results were averaged throughout all prediction times and manoeuvres}
	\label{fig:pred8}
\end{figure}

Previously obtained results for individual models open the possibility that by including training data from different test subjects with diverse enough driving styles, manoeuvre identification and prediction can be achieved for all test subjects.

\begin{figure*}[!ht]
	\centering

	\includegraphics[clip,width=1.7\columnwidth]{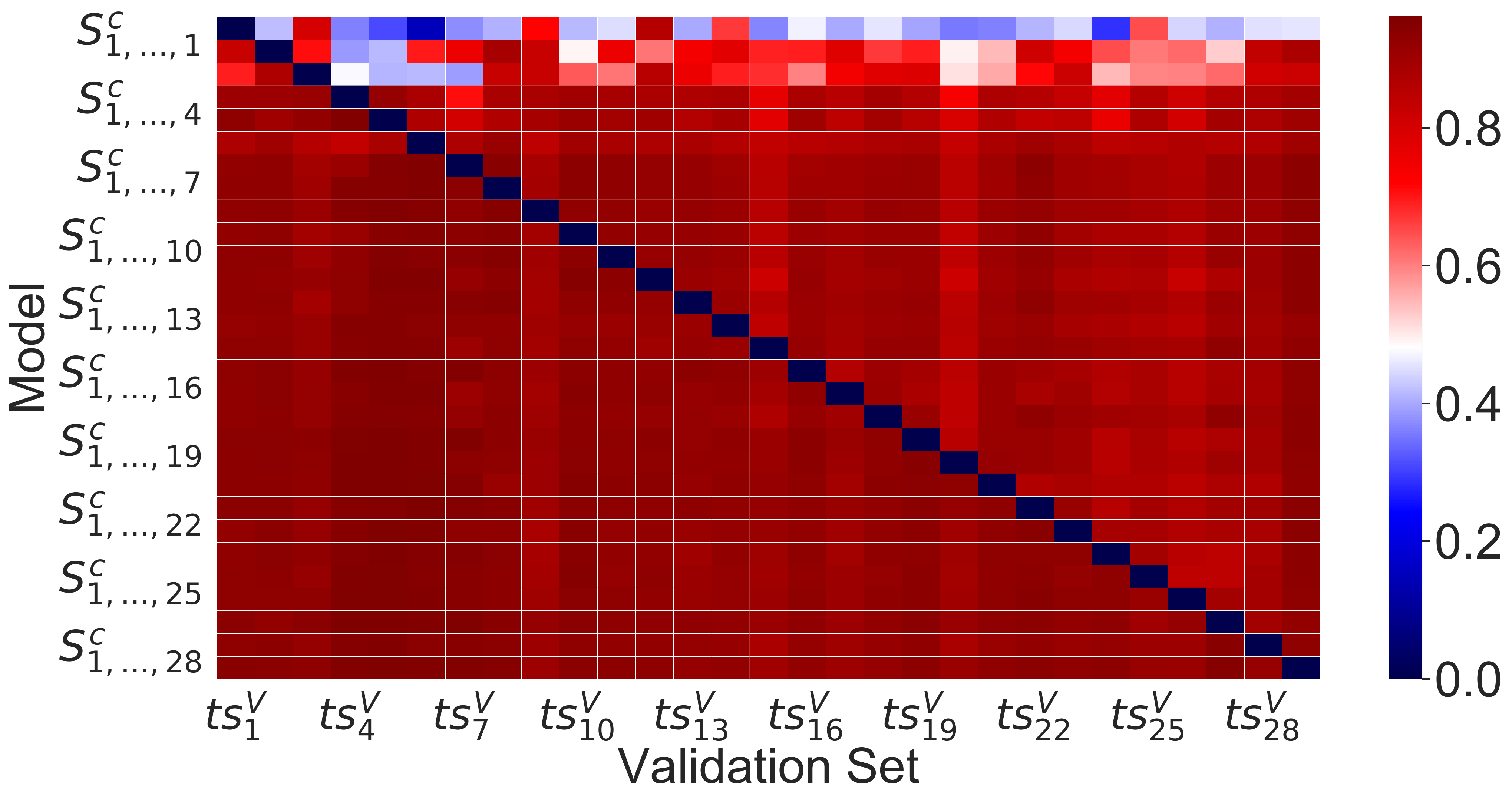}
	\caption{Mean F1 metric for models $S^{C}_{1,\dotsc,i}$ tested against test subjects $ts_{a}^{V}$ for all $i=\{1,2,\dotsc,N\}; a=\{1,2,\dotsc,N\}; i \neq a$. Results were averaged over all prediction times and manoeuvres}
	\label{fig:pred9}
\end{figure*}

\subsubsection{Test 3}
\label{sec:mm_3}
Predictions for groups of test subjects were tested as explained in Section \ref{sec:mttm}. Concatenated models (i.e. models trained with data from more than one test subject $S^{C}_{a,\dotsc,b}$) tested against known test subject $ts_{a,\dotsc,b}^{V}$. Figure \ref{fig:pred8} shows the prediction performance of all concatenated models for the same data set of test subjects; performance was measured using F1 metrics averaged by prediction window and manoeuvre. Overall performance was above $90\%$ for all models from $S^{C}_{1,\dotsc,7}$ onwards, an improvement against results from individual models in Section \ref{sec:mm_1} and Figure \ref{fig:pred6} which scored only above $80\%$. An interesting change in performance was shown when comparing models $S^{C}_{1}$, $S^{C}_{1,2}$ and $S^{C}_{1,2,3}$; performance drops in models $S^{C}_{1,2}$ onwards as data from different test subjects were added and generalization capability improves, with a sudden increase for both $S^{C}_{1,\dotsc,4}$ and $S^{C}_{1,\dotsc,5}$, followed by another drop for $S^{C}_{1,\dotsc,6}$. Behaviour before model $S^{C}_{1,\dotsc,4}$ could be explained by the intra-individual differences and highly limited data available for each test subject (i.e. model fits very different driving styles and performance drops as not enough data were available for these new styles) and behaviour of model $S^{C}_{1,\dotsc,6}$ could be linked to inter-individual differences of $ts_{6}$ that makes it especially difficult to model and generalize to (e.g. $S_{6}$ was the worst-performing model in Section \ref{sec:mm_1}, seen in Figure \ref{fig:pred6}). Models from $S^{C}_{1,\dotsc,7}$ onwards show good performance and generalization capability despite more test subjects being predicted.

These results led us to conclude that data from multiple test subjects can be used to train a model that can predict driving manoeuvres reliably among the known test subjects.

\subsubsection{Test 4}
\label{sec:mm_4}
Predictability of unknown test subjects was tested as explained in Section \ref{sec:mttm}. Concatenated models generalizing to predict behaviour of unknown test subjects was tested. Figure \ref{fig:pred9} shows prediction performance for concatenated models $S^{C}_{a,\dotsc,b}$ tested against all individual test subjects except $ts_{b}^{V}$; performance was measured using F1 metrics averaged by prediction window and manoeuvre. This tabular representation visualizes how different models perform when predicting different individual test subjects, with rows representing individual models and columns representing the test subject tested against. As hinted already by individual models’ generalization performance, by using training data from several test subjects, unknown test subjects can be reliably predicted; similar to the behaviour shown in Figure \ref{fig:pred8}, performance for models after $S^{C}_{1,\dotsc,4}$ considerably increases for unknown test subjects, with model $S^{C}_{1,\dotsc,10}$ already performing around $90\%$.

Therefore, the benefits of using data from more than one test subject to train the models are clear. From the time an additional training set was added (i.e. $S^{C}_{1,\dotsc, 2}$), performance increases substantially for several test subjects (from 0.15 to 0.7 for $ts^{V}_{6}$ or from 0.46 to 0.88 for $ts^{V}_{29}$); performance generally increased as more test subjects were added, with slight performance drops in some cases of 0.1 and 0.2. These results are intuitive regarding data-driven techniques and driving styles; as more data become available for the model to learn from, better performance and generalization capabilities can be achieved. However, due to the different driving styles and the difficulty of predicting them, slight drops in performance cannot be excluded at the current stage.

\begin{figure}[!ht]
	\centering
	\includegraphics[clip,width=\linewidth]{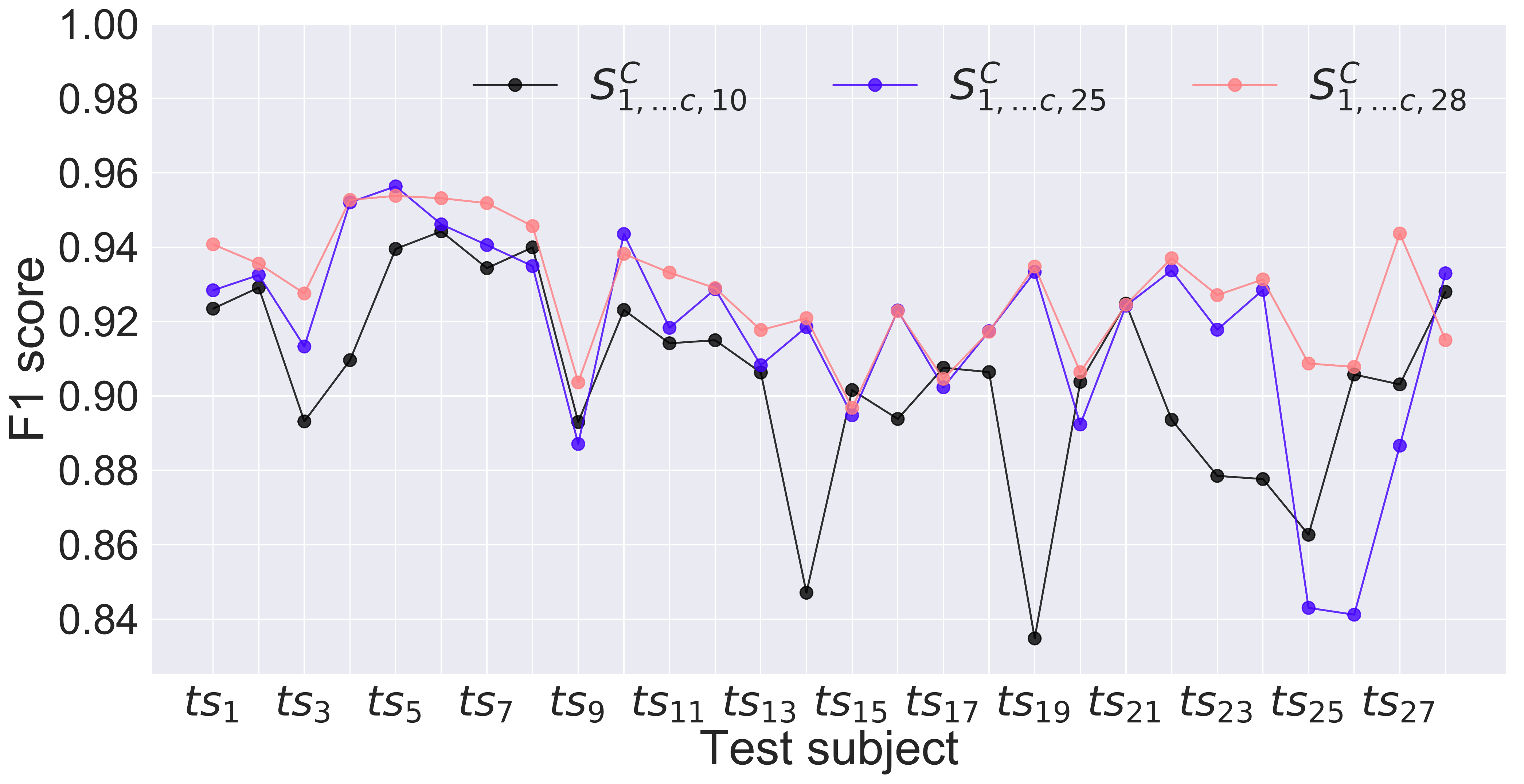}
	\caption{Mean F1 metric of models $S^{C}_{1,\dotsc,10}$, $S^{C}_{1,\dotsc,25}$ and $S^{C}_{1,\dotsc,29}$ tested against test subjects $ts_{i}^{V}$ for $i=\{1,2,\dotsc,N\}$. Results were averaged throughout all prediction times and manoeuvres}
	\label{fig:pred10}
\end{figure}

\begin{table*}[ht]
	\caption{Performance metrics statistics for manoeuvre identification ($t_{wo}=0$) and prediction ($t_{wo}=29$)}
	\label{tab:final3}
	\centering
	\begin{threeparttable}	
		\renewcommand{\arraystretch}{1.3}	
		\begin{tabular}{|c|c|c|c|c|c|c|c|c|}
			\cline{2-9}
			\multicolumn{1}{c|}{} & \multicolumn{4}{c|}{Identification} & \multicolumn{4}{c|}{Prediction} \\
			\hline
			F1 metric & ET & SVM & MLP & BI-LSTM & ET & SVM & MLP & BI-LSTM \\
			\hline
			Test 1 & \textbf{0.98} & 0.91 & 0.97 & 0.90 & 0.89 & 0.91 & 0.79 & \textbf{0.92} \\
			\hline
			Test 2 & 0.74 & 0.60 & 0.73 & \textbf{0.76} & - & - & - & \textbf{0.70} \\
			\hline
			Test 3 & \textbf{0.98} & 0.92 & 0.97 & 0.95 & $0.87^{*}$ & $0.90^{*}$ & $0.76^{*}$ & \textbf{0.92} \\
			\hline
			Test 4 & 0.87 & 0.86 & 0.90 & \textbf{0.93} & - & - & - & \textbf{0.82} \\
			\hline
		\end{tabular}
		\begin{tablenotes}
			\small
			\item[*] Only model $S_{1,\dotsc,29}^{V}$ is shown.
		\end{tablenotes}
	\end{threeparttable}
\end{table*}

Some additional analyses of the balance between generalization capability and performance whilst using the least amount of test subjects were performed to validate the results. General behaviour seen in Figure \ref{fig:pred9} is that performance tends to increase as more test subjects are added, with a performance for some test subjects decreasing: a model to model comparison of performance for individual test subjects could show a model that behaves close to optimal (i.e. $S^{C}_{1,\dotsc,29}$) but excluding some test subjects from the training phase to validate generalization capabilities. Figure \ref{fig:pred10} shows prediction performance of models $S^{C}_{1,\dotsc,10}$, $S^{C}_{1,\dotsc,25}$ and $S^{C}_{1,\dotsc,29}$ tested against all test subjects; performance was measured using F1 metrics averaged by prediction window and manoeuvre. $S^{C}_{1,\dotsc,25}$ shows performance close to optimal for $ts^{V}_{1,\dotsc,25}$ with loss on unknown test subject cases but staying close to 0.85, which makes it suitable for further analysis. An interesting behaviour seen in Figure \ref{fig:pred9} can be better visualized in Figure \ref{fig:pred10}, for the cases where performance did not increase as training data from more test subjects were added, showing initial decrease and a later increase (e.g. $ts^{V}_{9}$, $ts^{V}_{17}$, $ts^{V}_{20}$, $ts^{V}_{25}$, $ts^{V}_{26}$, $ts^{V}_{27}$); this behaviour points towards the impact training data from other test subjects might have on the model's performance (i.e. performance against a specific test subject could decrease with more data but improve if different test subjects were added). This mirrors a phenomenon seen in data-driven methods were making a dataset larger can negatively affect the performance of some models \cite{nakkiran2019deep} but increase it if augmented further; this phenomenon is still an open question in the field of data-driven methods but also validates our results from a numeric point of view.

\begin{figure}[!ht]
	\centering
	\includegraphics[clip,width=\columnwidth]{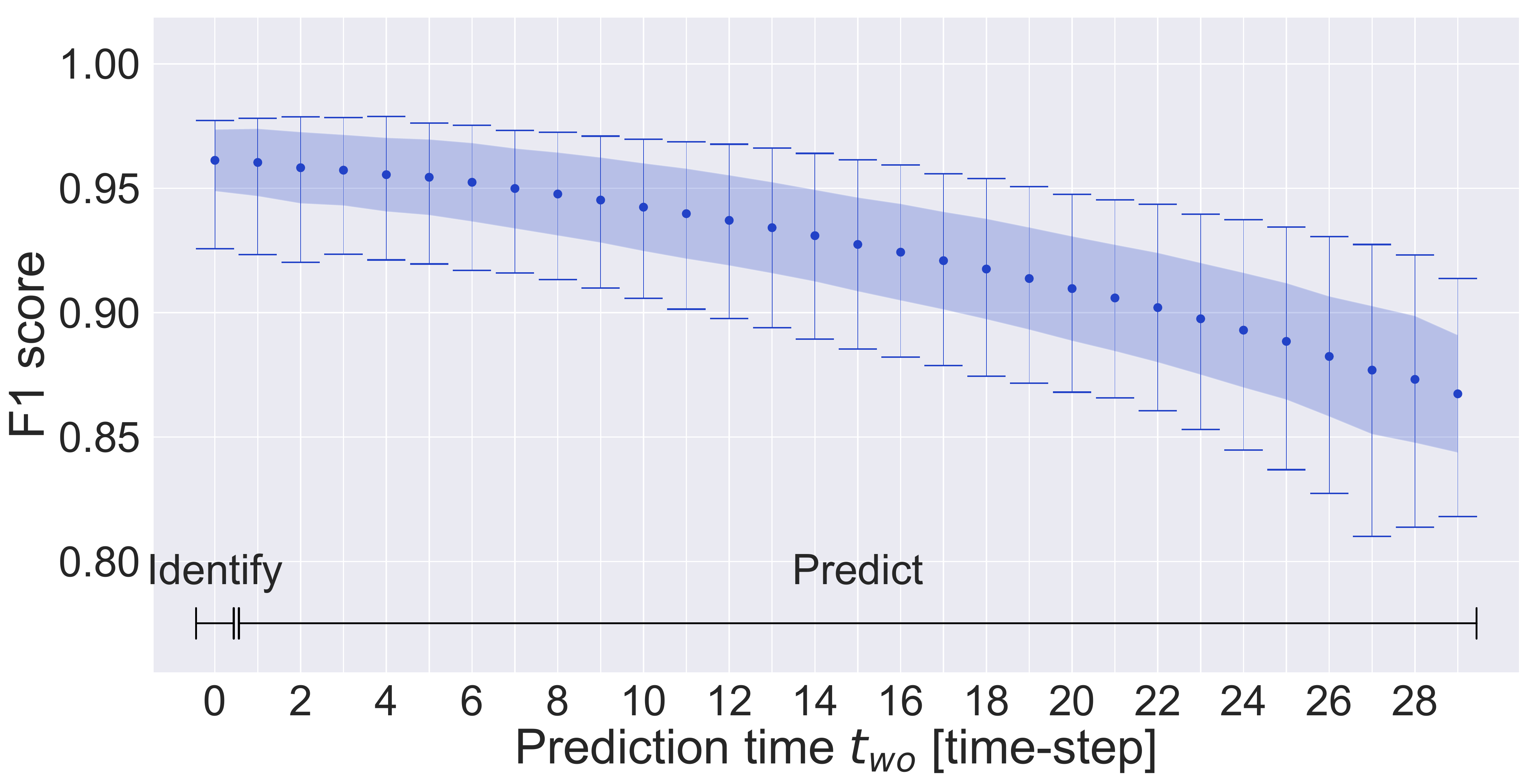}
	\caption{Statistical analysis of F1 metrics for model $S^{C}_{1,\dotsc,25}$ tested against test subjects $ts_{i}^{V}$ for $i=\{1,2,\dotsc,N\}$. Average of all $i$ for each prediction time (dot), standard deviation (shaded area), max and min values (caps)}
	\label{fig:pred11}
\end{figure}

Figure \ref{fig:pred11} shows prediction performance per $t_{wo}$ of $S^{C}_{1,\dotsc,25}$ tested against all test subjects; performance was measured using F1 metrics averaged by manoeuvres and among all test subjects; maximum and minimum values, as well as the standard deviation of the averaged results are displayed. The model performed above 0.80 for all combinations and $t_{wo}$ was better than individual models previously tested (see Figure \ref{fig:pred5}); it revealed considerable improvement for both manoeuvre identification (min 0.92, max 0.97, mean 0.96) and manoeuvre prediction (min 0.82, max 0.92, mean 0.86).

These results confirm that manoeuvres from known and unknown test subjects can be identified and predicted inside a certain prediction window, achieving a balance between precision and generalization capabilities.

\subsection{Comparative analysis between different methods}

Table \ref{tab:final3} shows the results for all 4 tests performed on the proposed model (i.e. BI-LSTM) compared to methods using shallow machine learning algorithms explored in previous work. Results were averaged across all test subjects. For manoeuvre identification, although current results did not outperform the previous ones for known test subjects (i.e. test 1 and 3), the final model outperformed older models for generalization to unknown test subjects (i.e. test 2 and 4). For manoeuvre prediction, current results outperformed previous results in every test; previous results failed to predict manoeuvres for unknown test subjects using single (i.e. test 2) or concatenated (i.e. test 4) models; for concatenated models, the best performing model for prediction using a shallow learning model was using data from all available test subjects $S_{1,\dotsc,29}^{V}$ in contrast to BI-LSTM models that could achieve performance between 0.80 and 0.90 for $t_{wo}=29$ without using data from all available test subjects. Additionally, the use of shallow learning models was only feasible when including the currently identified manoeuvre as input feature and after applying a re-sampling (under-sample the largest class and oversample the smallest classes) and relabelling process (create pre-manoeuvre classes of $n$ time-steps before a manoeuvre is performed) to the dataset.

An additional advantage achieved by the proposed model is the prediction granularity (i.e. predicting each time-step separately), which makes it flexible and easy to implement in different scenarios. Predictions per time-step with distinctive precision can be used to create confidence levels, either selectable by users in an ADAS recommender system, or to create an uncertainty bound over predictions integrated to an automatic controller.

Note that, although performance and generalization-wise BI-LSTM outperformed previous models based on shallow learning, the required training and testing times for the latter were considerably smaller (e.g. training of ET-based model $S_{1,\dotsc,29}^{V}$ takes $1/10$-th of the time for the BI-LSTM-based model in the same not-optimized machine); further improvements to reduce the training and testing times of these models are still under development, such as the use of multiple GPUs or the adoption of more optimizable convolutional architectures for sequence modelling such as WaveNet \cite{oord_wavenet:_2016}.

\section{Conclusions}
\label{sec:conclusions}

A new structure for manoeuvre identification and prediction whilst driving based on recurrent neural networks and HRI principles was presented. By integrating human-based and car-based features as a time sequence, a model that links and associates them with driving manoeuvres was achieved. Current and future manoeuvre prediction was accomplished for known and unknown test subjects, achieving performance metrics around $90\%$. Integrating our model in a real vehicle could be achieved, in the form of an advanced ADAS that ensures driver safety; a GPS and roadmap-aware ADAS could compare current and predicted manoeuvres with expected safe behaviour according to the situation, and warn drivers using visual or audio cues. Furthermore, reducing a driver's control over the vehicle could be achieved in a cooperative driving scheme. Predictive turn signalling could also be explored for cases where turn signals were not raised by the driver.

These results open the possibility of using such methods for task identification and prediction for general HRI. For example, cooperative object handover \cite{parastegari_failure_2018} or object reaching \cite{mainprice_goal_2016} with robots tracking and predicting user's limb movement. Additional use for teleoperation is feasible for integration with predictive controller schemes \cite{sirouspour_model_2006}.

Future work will focus on integrating more human-based data into our models to enable earlier prediction. Additional experiments using a physical robotic-vehicle platform are planned to further validate our results.

\section*{Acknowledgment}
The authors want to thank the continuous support of the University of Bristol in technical aspects and COLCIENCIAS, the Newton Fund and EPSRC throughout APC Spoke Bath for funding this work as part of the lead author's PhD studies \cite{pulgarin2019estimation}.


\bibliographystyle{IEEEtran}
\bibliography{article}

\end{document}